\documentclass[10pt,twocolumn,letterpaper]{article}
\usepackage[pagenumbers]{cvpr}

\usepackage{graphicx}
\usepackage{amsmath}
\usepackage{amssymb}
\usepackage{booktabs}

\usepackage[utf8]{inputenc} 
\usepackage[T1]{fontenc}    
\usepackage{xcolor}         
\definecolor{citecolor}{HTML}{0071bc}
\usepackage[pagebackref,breaklinks=true,colorlinks,citecolor=citecolor,bookmarks=false]{hyperref}
\usepackage[capitalize]{cleveref}
\usepackage{url}            
\usepackage{booktabs}       
\usepackage{amsfonts}       
\usepackage{nicefrac}       
\usepackage{microtype}      
\usepackage{graphicx}
\usepackage{amsmath}
\usepackage{algorithm}
\usepackage{comment}
\usepackage[noend]{algpseudocode}
\usepackage{bbm}
\usepackage{dsfont}
\usepackage{easylist}
\usepackage{xspace}
\usepackage{wrapfig}
\usepackage{multirow} 
\usepackage{amssymb}

\usepackage{subcaption}
\newcommand{\myparagraph}[1]{\vspace{2pt}\noindent{\textbf{#1}}}

\DeclareMathOperator*{\argmax}{argmax}

\definecolor{skyblue1}{RGB}{114, 159, 207}
\definecolor{scarletred3}{RGB}{164,   0,   0}
\definecolor{mytangoblue}{RGB}{59, 109, 172}
\definecolor{mytangoorange}{RGB}{243, 121, 33}

\newcommand{\rowNumber}[1]{\textcolor{skyblue1}{#1}}

\definecolor{green4}{RGB}{  0, 139,   0}

\newcommand{\prop}{{\text{prop}}}
\newcommand{\init}{{\text{init}}}

\newcommand{\nmsdets}{{NMS-based detectors\xspace}}

\newcommand{\ddetr}{{\textit{Deformable-DETR}}\xspace}

\newcommand{\reffig}[1]{Figure~\ref{fig:#1}}

\newcommand{\refsec}[1]{Section~\ref{sec:#1}}

\newcommand{\reftbl}[1]{Table~\ref{tbl:#1}}

\newcommand{\refeq}[1]{Equation~\eqref{eq:#1}}
\newcommand{\refeqshort}[1]{\eqref{eq:#1}}

\newcommand{\lblfig}[1]{\label{fig:#1}}
\newcommand{\lblsec}[1]{\label{sec:#1}}
\newcommand{\lbleq}[1]{\label{eq:#1}}

\newcommand{\lbltbl}[1]{\label{tbl:#1}}

\makeatletter
\DeclareRobustCommand\onedot{\futurelet\@let@token\@onedot}
\def\@onedot{\ifx\@let@token.\else.\null\fi\xspace}
\def\etal{\emph{et al}\onedot}

\def\name{{\textit{DETA}}\xspace}

\def\highlighttitle{\textbf{De}tection \textbf{T}ransformers with \textbf{A}ssignment \xspace}

\crefname{section}{Sec.}{Secs.}
\Crefname{section}{Section}{Sections}
\Crefname{table}{Table}{Tables}
\crefname{table}{Tab.}{Tabs.}

\begin{document}
\title{NMS Strikes Back}

\author{Jeffrey Ouyang-Zhang \quad
Jang Hyun Cho \quad
Xingyi Zhou \quad 
Philipp Kr\"ahenb\"uhl \vspace{3mm}\\
The University of Texas at Austin
}
\maketitle

\begin{abstract}
Detection Transformer (DETR) directly transforms queries to unique objects by using one-to-one bipartite matching during training and enables end-to-end object detection.
Recently, these models have surpassed traditional detectors on COCO with undeniable elegance.
However, they differ from traditional detectors in multiple designs,
including model architecture and training schedules,
and thus the effectiveness of one-to-one matching is not fully understood.
In this work, we conduct a strict comparison between the one-to-one Hungarian matching in DETRs and the one-to-many label assignments in traditional detectors with non-maximum supervision (NMS).
Surprisingly, we observe one-to-many assignments with NMS consistently outperform standard one-to-one matching under the same setting, with a significant gain of up to $2.5$ mAP.
Our detector that trains \ddetr{} with traditional IoU-based label assignment achieved $50.2$ COCO mAP within 12 epochs ($1\times$ schedule) with ResNet50 backbone, outperforming all existing traditional or transformer-based detectors in this setting.
On multiple datasets, schedules, and architectures, we consistently show bipartite matching is unnecessary for performant detection transformers.
Furthermore, we attribute the success of detection transformers to their expressive transformer architecture.
Code is available at \href{https://github.com/jozhang97/DETA}{https://github.com/jozhang97/DETA}.
\end{abstract}

\section{Introduction}

Traditional object detectors~\cite{ren2015faster,lin2017focal} treat detection as region (two-stage) or anchor (one-stage) classification.
They assign the ground truth labels to one or many regions/ anchors and produce a set of overlapping predictions for each object.
They then remove duplicates via post-processing, namely non-maximal suppression (NMS).
Recent transformer-based detectors~\cite{carion2020end} side-step explicit NMS, and learn a suppression strategy in training via a strict one-to-one bipartite matching loss.
Over two years of exciting development, end-to-end detectors have shown significant progress~\cite{dai2021up,zhu2020deformable,dai2021dynamic,wang2021anchor,cheng2021maskformer,sun2020rethinking,yao2021efficient,meng2021conditional,liu2022dab,li2022dn} and recently surpassed the performance of best traditional detectors~\cite{zhang2022dino}.
However, detection transformers differ from traditional detectors along various axes: end-to-end design with Hungarian matching, heavier architectures (12 attention layers vs. 2 linear layers), longer training schedules (50 -- 500 epochs vs. 12 epochs), etc.
This makes it challenging to pinpoint what exactly contributes to good overall performance.

\begin{figure}
    \centering
    \begin{subfigure}{0.49\columnwidth}
        \includegraphics[width=1.0\columnwidth]{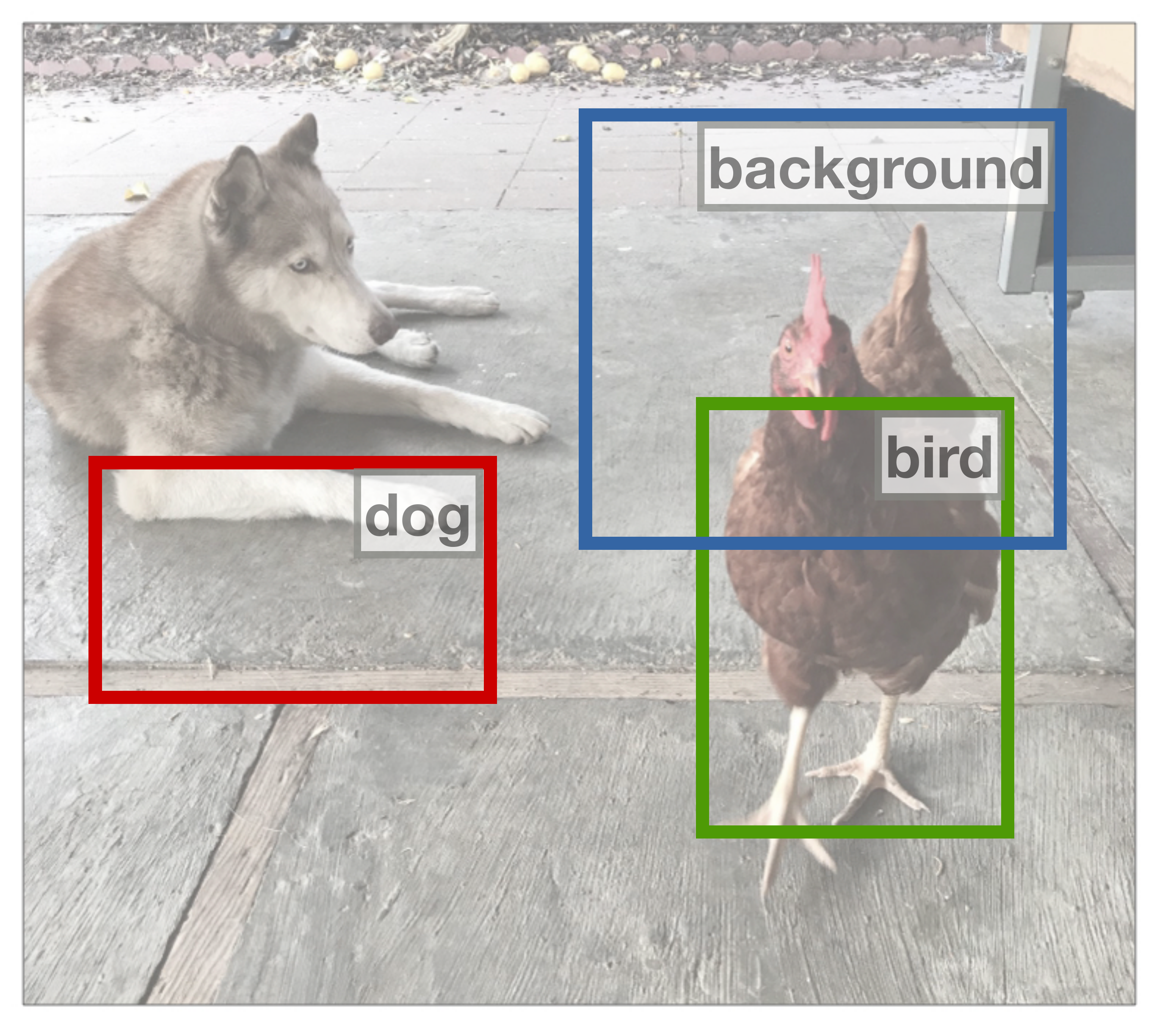}
        \caption{DETR's Bipartite matching}
    \end{subfigure}
    \begin{subfigure}{0.49\columnwidth}
        \includegraphics[width=1.0\columnwidth]{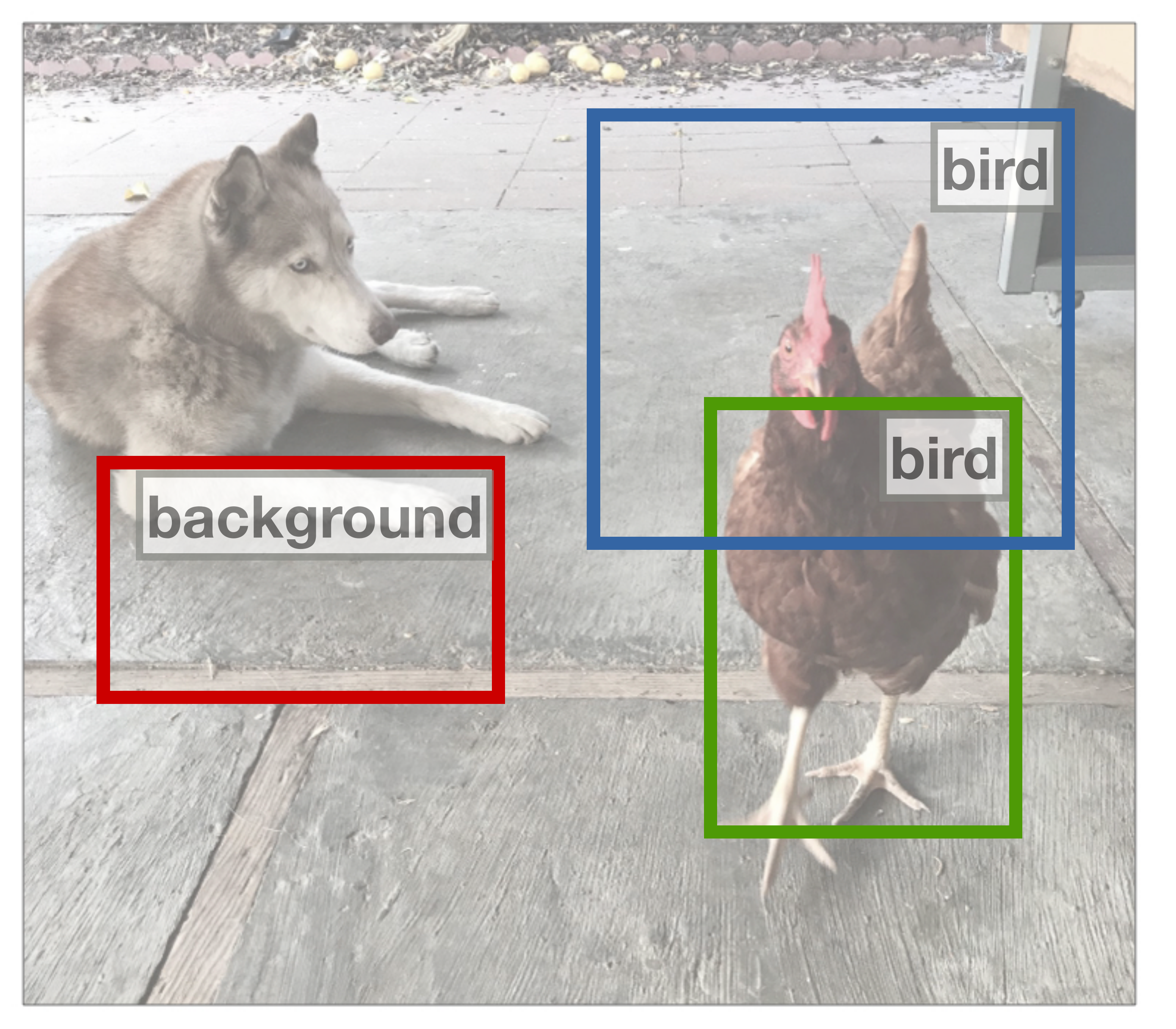}
        \caption{Traditional IoU assignment (ours)}
    \end{subfigure}
    \caption{{\textbf{Detection Transformers vs. with traditional training objective.}
    Presumable predicted boxes are visualized with ground truth object labels.
    (a) DETR-style bipartite matching matches one model prediction to one ground truth object such that the matched losses are minimized.
    (b) traditional IoU assignment assigns multiple anchors/proposals with high IoU to each ground truth object.
    Multiple positive labels enrich the training signal.
    }}
    \lblfig{matching}
\end{figure}

In this paper, we correct a widely-believed misconception that one-to-one mapping is essential for highly-performant detection.
\textit{Rather, the traditional one-to-many training objective yields equally proficient detection transformers.}
We design a transformer-based object detector that directly \emph{assigns} positive-negative labels to each query as in traditional detectors~\cite{ren2015faster,lin2017focal} 
and use NMS to remove duplicated predictions, instead of using end-to-end one-to-one \emph{matching}.
We name our detector \name (\highlighttitle).
\footnote{Here we highlight the distinction between ``matching'' and ``assignment'': ``matching'' is from a dynamic algorithm, e.g., Hungarian matching; ``assignment'' is from a fixed rule, e.g., IoU-based assignment.}
Specifically, we follow the strong two-stage DETR framework~\cite{zhu2020deformable},
and replace the one-to-one Hungarian matching loss with a one-to-many IoU-based assignment loss in both stages.
In the first stage, we assign anchors from a transformer encoder to annotations or background using an IoU metric.
In the second stage, we assign queries similarly. %
Our detector produces overlapping objects and uses NMS for post-processing.
\name{} has precisely the same architecture as its end-to-end counterparts. %

Our hybrid architecture allows us to conduct strict comparisons with end-to-end transformer detectors and traditional detectors.
Compared to traditional detectors~\cite{ren2015faster,lin2017focal}, \name features a more expressive head:
In the first stage, \name uses a 6-layer transformer encoder over the convolutional backbone, while a traditional RPN~\cite{ren2015faster} or RetinaNet~\cite{lin2017focal} uses 2 to 4 convolutional layers;
in the second-stage, \name uses a 6-layer transformer decoder, while a traditional FasterRCNN~\cite{ren2015faster} uses 2 linear layers.
Compared to end-to-end transformer detectors~\cite{carion2020end,zhu2020deformable}, \name only differs in training loss, with exactly the same architecture.
The effectiveness of our training loss comes from two properties.
First, one-to-many mapping yields more total positive predictions which help speed up model convergence.
Second, we use an object balancing technique that helps stabilize matching for small objects, which are shown to be especially challenging in end-to-end detectors~\cite{carion2020end}.
We verify the advantages of both properties in our controlled experiments.

Our resulting detector is effective and easy to train.
We show switching from standard bipartite matching to our IoU assignments in Deformable DETR~\cite{zhu2020deformable} brings a significant $2.5$ mAP improvement.
Our model converges much faster than the existing transformer-based detector with superior performance, with $50.2$ mAP on COCO under the standard 1$\times$ schedule (12 epochs) and ResNet50 backbone.
This is $7.3$ points higher than the best NMS-based detector ($42.9$ mAP of CenterNet2~\cite{zhou2021probablistic}) and $0.8$ points better than a strong end-to-end detector ($49.4$ mAP of DINO~\cite{zhang2022dino}).
Our observation is consistent across different detection transformer architectures (DETR~\cite{carion2020end} and Deformable DETR~\cite{zhu2020deformable}) and datasets (COCO~\cite{lin2014coco} and LVIS~\cite{gupta2019lvis}).
On longer schedules and larger backbones, our traditional training objective can still replace bipartite objectives.

\section{Related Work}

\myparagraph{Traditional \nmsdets} map the input image to a downsampled dense feature map.
Each pixel in the feature map is classified into an object class or background, resulting in overlapping predictions for each object.
RetinaNet~\cite{lin2017focal} and YOLO~\cite{redmon2016you,redmon2017yolo9000,redmon2018yolov3} define object or background by predefined IoU thresholds.
CenterNet~\cite{zhou2019objects}, FCOS~\cite{tian2019fcos}, ATSS~\cite{zhang2020bridging}, and GFL~\cite{li2020generalized} improve the IoU-based positive-negative definition to adaptive criteria.
While multiple positive selection strategies~\cite{zhang2020bridging,chen2021you} are compatible with our method, we utilize the most common and straightforward fixed IoU threshold.
Optionally, NMS-based detectors can be improved by a second stage~\cite{ren2015faster,he2017mask,cai2018cascade,chen2019hybrid,zhou2021probablistic}, 
where the region features are refined using the IoU thresholds.
Again, the second stage outputs are duplicated and need NMS for post-processing.

Our work uses the simple IoU-based criterion from traditional detectors but also a transformer architecture.

\myparagraph{End-to-end detectors}.
DETR~\cite{carion2020end} removes the duplication in detectors' predictions by directly mapping queries to unique objects using self-attention.
The vanilla DETR is hard to train and has been improved by multiple follow-up works.
Deformable-DETR~\cite{zhu2020deformable} introduces multi-scale features with a deformable operation.
Deformable-DETR, TSP-RCNN~\cite{sun2020rethinking}, and EfficientDETR~\cite{yao2021efficient} identify the learned object query as a causal factor for slow-convergence and replace them with a first-stage detector.
Conditional-DETR~\cite{meng2021conditional} and DAB-DETR~\cite{liu2022dab} concretize the implicit query as explicit point locations or boxes.
DN-DETR~\cite{li2022dn} eases the one-to-one matching task by adding an auxiliary query de-noise loss.
DINO~\cite{zhang2022dino} combines 2-stage queries and learned queries, and finally surpasses the performance of any existing \nmsdets.

All these works bring partial wisdom from \nmsdets{} but keep the Hungarian matching losses untouched.
Our work is complementary to these ideas - we show a simple change in matching provides significant improvements.

GroupDETR~\cite{chen2022group} and $\mathcal{H}$-DETR~\cite{jia2022detrs} concurrently identified one-to-one matching as a training bottleneck.
They train with additional queries in the transformer decoder to increase the number of matched queries and observe faster convergence and better performance.
This validates our insight that more positive samples help.
FQDet~\cite{picron2022fqdet} finds that techniques from traditional detectors, such as IoU assignment and multiple anchor ratios, improve the convergence of detection transformers, however, they did not show significant performance improvements over detection transformers as ours.
Our detector does not modify the architecture except for adding NMS and only modifies the training loss. 
We compare with these approaches in Sections \ref{sec:prior_works} and \ref{sec:larger_backbones}.

\begin{figure*}
    \centering
    \includegraphics[width=2.0\columnwidth]{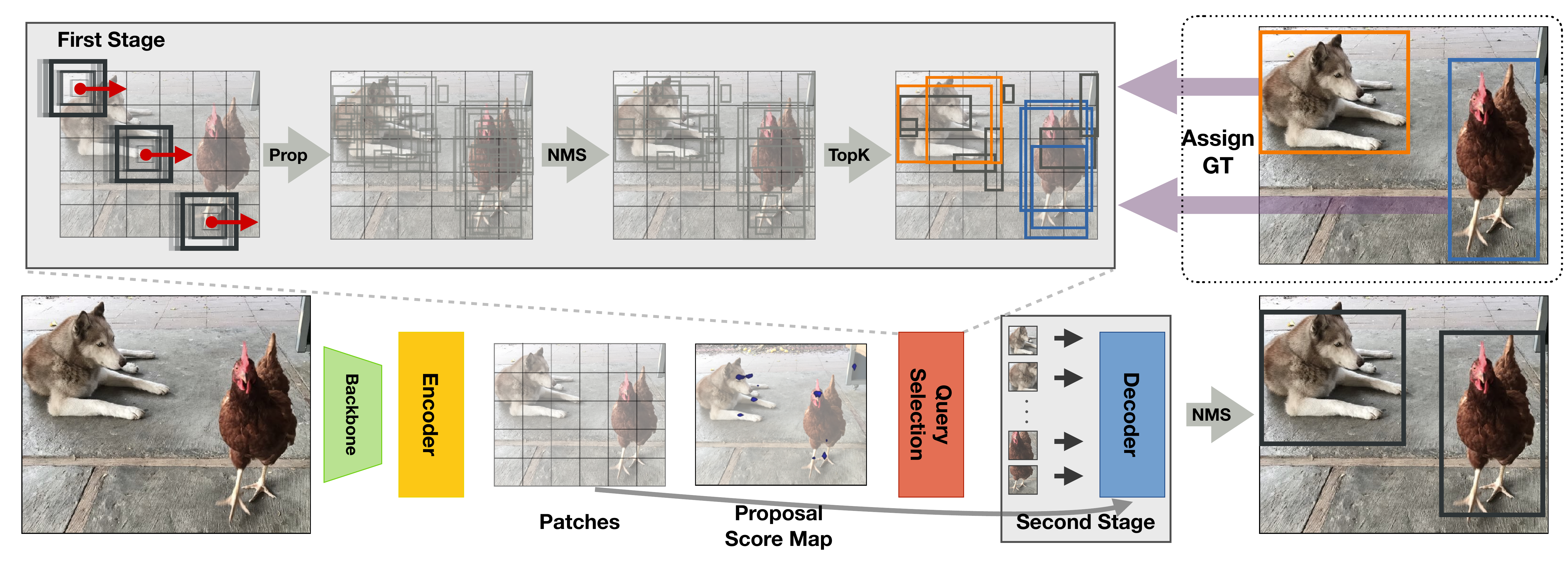}
    \caption{
    \textbf{Overview of our detection framework.}
    The transformer encoder extracts image features and a proposal score map (Bottom left).
    We sample top-ranking boxes as queries (Top left) and decode them with the image features.
    Each query reads out classification logits and a bounding box.
    The dotted section (Top right) shows our assignment procedure for the second stage during training.
    The \textcolor{mytangoorange}{orange} and \textcolor{mytangoblue}{blue} boxes indicate assigned proposals and objects.
    }
    \lblfig{pipeline}
\end{figure*}
\section{Preliminary}

Given an image $I\in\mathbb{R}^{H\times W\times 3}$, a detector produces a bounding box $b_i\in\mathbb{R}^4$ and scores $s_i\in\mathbb{R}^{C}$ for $C$ categories for each object $i$.

\myparagraph{DETR}~\cite{carion2020end} decomposes detection into three components: a backbone, a transformer encoder, and a transformer decoder.
The backbone takes the image as an input and produces a down-sampled $D$-dimensional feature map $F\in\mathbb{R}^{M\times D}$.
Here $M$ refers to the number of features in the feature map,
$M=w \times h$ in vanilla DETR~\cite{carion2020end} or $M=\sum_l w_l \times h_l$ in multi-scale feature resolutions~\cite{zhu2020deformable}.
The \emph{transformer encoder} refines $F$ using self-attention with positional embeddings and produces updated features $F^\prime\in\mathbb{R}^{M \times D^\prime}$.
The \emph{transformer decoder} transform a set of queries $Q\in \mathbb{R}^{N\times D^\prime}$ into $N$ unique object detections (or background) $O = \{\hat{b}_i, \hat{s}_i\}_{i=1}^N$.
During this process each query cross-attends to the encoded feature map $F^\prime$.
Self-attention layers ensure objects $O$ have little to no duplicates, and thus can learn non-maximum-suppression implicitly.

\myparagraph{Two-stage DETR.} In the vanilla DETR~\cite{carion2020end}, the object queries $Q$ are learnable network parameters and are fixed for all testing images.
Multiple works~\cite{zhu2020deformable,sun2020rethinking,yao2021efficient,zhang2022dino} show image-dependent queries improve convergence and performance.
In the two-stage Deformable-DETR~\cite{zhu2020deformable}, the transformer encoder densely predicts ranked proposal boxes $b^{\prop}_i$ from fixed initial boxes $b^{\init}_i$.
The top N proposals form queries for a second stage $Q_i = \mathcal{G}(b^{\prop}_i)$.
The second stage now takes features from the first stage as inputs instead of fixed queries.
We follow this two-stage query formulation.

\myparagraph{Hungarian matching loss.}
During training, DETR assigns each network output $O=\{\hat b_j, \hat s_j\}_{j=1}^{M}$ to exactly one annotated object in $G=\{b_k, c_k\}_{k=1}^{K}$ or background $\emptyset$ where $c_k$ is the class id for object $k$.
Let $\sigma_i \in \{\emptyset, 1 \ldots K\}$ be the assignment of network output $i$ to annotation $\sigma_i$.
DETR enforces a one-to-one matching to learn non-maxima-suppression: For $i \ne j$ with $\sigma_i\ne\emptyset$ and $\sigma_j\ne\emptyset$, we have $\sigma_i \ne \sigma_j$.
Furthermore for each annotation $k$, $\exists i : \sigma_i=k$.
DETR optimizes for the best matching in the training objective:

\begin{equation}
 \mathcal{L}(O, G) = \min_\sigma\underbrace{
 \left( \sum_{i=1}^M \mathcal{L}_{cls}(\hat s_i, c_{\sigma_i}) + \mathcal{L}_{box}(\hat b_i, b_{\sigma_i})\right)}
 _{\mathcal{L}(O, G | \sigma)}
 \lbleq{matching}
\end{equation}

where $\mathcal{L}_{cls}$ is a focal-loss based classification loss, and $\mathcal{L}_{box}$ is a bounding box loss composed of GIoU and L1 loss. 
Any unmatched output $\sigma_i = \emptyset$ predicts a background score $s_{\emptyset}$, and incurs no box loss $\mathcal{L}_{box}(\hat b_i, b_\emptyset) = 0$.
DETR finds the optimal matching $\sigma$ using the Hungarian algorithm
\footnote{In practice, DETR uses a different cost to match and supervise training.}.

In the following section, we show we can replace the one-to-one matching with a simple overlap-based assignment strategy to significantly boost the performance and training speed of DETR.

\section{IoU assignment in transformer detectors}

\name changes the assignment procedure in \refeq{matching}.
We closely follow the assignment strategies of traditional two-stage detectors~\cite{ren2015faster}.
First, we relax the one-to-one matching constraint and instead allow one-to-many assignments.
We allow multiple outputs $i$ and $j$ to be assigned to the same ground truth $\sigma_i = \sigma_j$.
Second, we switch to a \emph{fixed} overlap-based assignment strategy instead of the loss-based matching in \refeq{matching}.

\myparagraph{First stage assignments.}
We start from the fixed initial query features $F$ derived from the Deformable-DETR architecture.
Each query $i$ is anchored at a specific location $(x_i,y_i)$ within the image.
However, proposals do not have a natural width and height required for box overlap metrics.
We use a constant box size $w_i = 0.1 \times 2^{-l_i} \times W$ and $h_i = 0.1 \times 2^{-l_i} \times H$, where $l_i \in \{0, \ldots, 3\}$ is the feature resolution, $W$ and $H$ are the width and height of the image.
This yields an initial box definition $b^\init_i = (x_i -\frac{w_i}{2}, y_i -\frac{h_i}{2}, x_i +\frac{w_i}{2}, y_i +\frac{h_i}{2})$.
We use these initial boxes for both positional embeddings and the overlap-based assignment process.
We assign each prediction $i$ to the highest overlapping ground truth object if the overlap is larger than a threshold $\tau$.

At the beginning of training, a ground truth object sometimes does not have an overlapping prediction with a threshold $\tau$.
We find no difference whether we keep the object unmatched or assigned to the closest unmatched prediction.
For mathematical convenience here, we assume the latter.
Anchor $i$ is closest to object $k$ if the condition
\begin{equation}
C^{\init}_{\max}(i,k)=\bigcap_{j\neq i} \mathbbm{1} \{ IoU(b_i^{\init}, b_k) \ge IoU(b_j^{\init}, b_k) \}
\end{equation}
is met. Then, the assignment procedure is formally
\begin{equation}
\small
    \sigma^{\init}_i = 
\begin{cases}
    \hat k = \argmax_k IoU(b_i^{\init}, b_k),
    & \text{if } IoU(b_i^{\init}, b_{\hat k}) \ge \tau\\
    & \text{or } C^{\init}_{\max}(i,\hat k) \\
    \emptyset,                & \text{otherwise}
\end{cases}
\lbleq{assignment}
\normalsize
\end{equation}
We use a threshold $\tau=0.7$ for the first stage.
Since the first stage uses a single box width and height per query feature, the resulting overlap metric $IoU$ is very similar to the centerness estimate in anchor-free detectors~\cite{zhou2021probablistic,tian2019fcos,zhou2019objects}.

We train the first stage using a class-agnostic version of the DETR loss $\mathcal{L}(O, G | \sigma^{\init})$ \refeqshort{matching} on our fixed assignments $\sigma^{\init}$ \refeqshort{assignment}.
The classification loss $\mathcal{L}_{cls}$ of the first stage is a binary focal loss indicating foreground vs background.
The first stage assignment procedure and loss formulation naturally encourage (near-)duplicate outputs.
We follow Ren~\etal~\cite{ren2015faster} and apply non-maxima-suppression on the proposal boxes $b^\prop$.
It reduces the computational burden and increases coverage of the first-stage proposals.

\myparagraph{Second stage assignments.}
The assignment procedure in the second stage closely follows the first stage assignment procedure.
However, instead of using a fixed set of initial boxes $b^\init$, we use the outputs of the first stage $b^\prop$ to define an equivalent assignment $\sigma^\prop_i$.
We train this second stage using the same loss as DETR $\mathcal{L}(O, G|\sigma^\prop)$.
Following Faster RCNN~\cite{ren2015faster}, we balance the foreground object ratio using a hyper-parameter $\gamma$.
Specifically, when the number of positive queries is larger than $N \times \gamma$, we randomly sample $N \times \gamma$ positive queries from them.
To encourage a larger number of positive queries we relax the overlap criterion for the second stage to $\tau=0.6$.
This simple relaxation of the overlap criterion already works well for most mid-sized and large objects.
However, small objects are under-sampled.

\begin{table*}[ht]
\centering
\newcommand{\betternum}[2]{#1\textcolor{green4}{\scriptsize{(+#2)}}}
\newcommand{\nicenum}[2]{\textbf{#1}\textcolor{green4}{\scriptsize{(+#2)}}}
\begin{tabular}{@{}l@{\ \ }l@{\ \ \ }c@{\ \ \ }c@{\ \ \ }c@{\ \ \ }c@{\ \ \ }c@{\ \ \ }c@{\ \ \ }c@{\ \ \ }c@{\ \ \ }c@{\ \ \ }c@{\ \ \ }c@{\ \ \ }c@{}}
\toprule
              &                            & \multicolumn{8}{c}{COCO}                                            & \multicolumn{4}{c}{LVIS}                           \\          
\cmidrule(r){3-10}
\cmidrule(r){11-14}
              &                            & \multicolumn{4}{c}{\ddetr{}~\cite{zhu2020deformable}}   & \multicolumn{4}{c}{DETR~\cite{carion2020end}}     & \multicolumn{4}{c}{\ddetr{}~\cite{zhu2020deformable}}                                             \\
\rowNumber{\#} &                            & $AP$                   & $AP_s$ & $AP_m$ & $AP_l$  & $AP$                & $AP_s$ & $AP_m$ & $AP_l$ & $AP$                    & $AP_s$ & $AP_m$ & $AP_l$ \\
\cmidrule(r){1-2}
\cmidrule(r){3-6}
\cmidrule(r){7-10}
\cmidrule(r){11-14}
\rowNumber{1}  & \textcolor{gray}{Baseline} & \textcolor{gray}{43.2} & \textcolor{gray}{25.9}   & \textcolor{gray}{46.6}   & \textcolor{gray}{57.6}   & \textcolor{gray}{35.3} & \textcolor{gray}{15.2} & \textcolor{gray}{37.5} & \textcolor{gray}{53.6} & -                   & -      & -      & -       \\
\rowNumber{2}  & Improved baseline                     & 47.7    & 31.2   & 51.0   & 61.8   & 41.8          & 22.3 & 44.6 & 59.7  & 31.5  & 22.1   & 41.2   & 48.3    \\ 
\rowNumber{3}  & + IoU Asgmt \& NMS w/o Object balance & 48.7    & 30.9   & 53.1   & 64.4   & 41.8          & 21.3 & 46.2 & 61.2  & 31.5  & 21.5   & 41.0   & 50.0   \\
\rowNumber{4}  & + Object balance (Ours)      & \textbf{50.2}    & 32.2   & 54.4   & 64.6   & \textbf{43.9} & 24.6 & 48.2 & 60.8  & \textbf{33.9} & 23.2   & 44.0   & 52.7    \\
\bottomrule
\end{tabular}
\caption{
\textbf{IoU assignments vs. Hungarian matching.}
We report AP on COCO ($1\times$ schedule for \ddetr{}, $50$ epoch for DETR) and LVIS (180k iter) validation with ResNet50 backbone.
Row \rowNumber{1}: the original baselines;
Row \rowNumber{2}: our improved baseline; %
Row \rowNumber{3}: changing Hungarian matching in Row \rowNumber{2} to naive IoU assignment and re-introducing NMS;
Row \rowNumber{4}: applying our object balancing technique (\refsec{balance}) to Row \rowNumber{3}.
IoU assignment with object balancing significantly outperforms Hungarian matching.
See \reftbl{r50} for longer schedules on COCO.
}
\lbltbl{ablation}
\end{table*}

\myparagraph{Object Balancing.}
\lblsec{balance}
In contrast to the one-to-one mapping that always has one matched prediction for each ground truth object,
our IoU assignment might assign a dramatically different number of predictions across ground truth objects.
Larger objects are naturally more likely to contain many overlapping predictions than smaller objects.
This imbalance hurts performance.
We propose a simple object-balancing technique that samples at most $n$ positive assignments for each ground truth.
Let $\mu^n_k$ be the $n$-th highest IoU for annotation $k$ and $\tau_k^n = \max(\tau, \mu^n_k)$ be our new dynamic threshold.
We modify the assignment procedure in \refeq{assignment}:

\begin{equation}
\small
    \sigma^{\prop}_i = 
\begin{cases}
    \hat k = \argmax_k IoU(b_i^{\prop}, b_k), & \text{if } IoU(b_i^{\prop}, b_{\hat k}) \ge \tau^n_k \\
    & \text{or } C^{\prop}_{\max}(i,\hat{k}) \\
    \emptyset,                & \text{otherwise}
\end{cases}
\lbleq{assignment_n}
\normalsize
\end{equation}
We verify balancing objects benefits small object performance in~\refsec{positives}.

\section{Results}
We evaluate our method on the COCO 2017 and LVIS benchmarks~\cite{lin2015microsoft,gupta2019lvis}.
COCO contains 118k training and 5k validation images over 80 object categories.
LVIS is a long-tail detection dataset with 1203 object categories.
We report bounding box mAP as the evaluation metric.

\subsection{Implementation Details}
\lblsec{details}
Unless other specified, we build off two-stage Deformable-DETR~\cite{zhu2020deformable}.
We follow detectron2~\cite{wu2019detectron2} to set traditional detection hyperparameters.
Specifically, we set the IoU threshold $\tau'=0.7$ in the first stage and $\tau=0.6$ in the second stage.
The foreground ratio $\gamma$ is $0.5,0.25$ in the first and second stages, respectively.
We use an NMS threshold of $0.9,0.7$ in the first and second stage.
Prior to NMS, we filter to 1000 boxes per feature level in the first stage and 10000 boxes in the second stage.
We set the object-balancing parameter $K=4$ (\refsec{balance}) as swept in \reftbl{balance}.

\myparagraph{Improved DETR and Deformable-DETR baselines.}
For our main results, we start with an improved two-stage Deformable-DETR~\cite{zhu2020deformable} baseline following DINO~\cite{zhang2022dino} with 900 queries (or proposals), 5 feature scales, larger feed-forward dimension, box refinement, etc.
We also report performance with vanilla DETR by replacing deformable attention with vanilla attention.
Due to memory constraints in vanilla self-attention, DETR uses features $F$ from the last stage of the backbone with extra dilation and reduced stride (typically known as DC-5).
The number of proposals becomes much smaller - for a $480 \times 480$ image, $M=900$ for DETR's DC-5 features whereas $M\approx 4800$ for Deformable-DETR's multi-scale features.
Since the number of proposals is close to the number of queries, we use 300 queries and keep bipartite matching in the first stage of DETR and only use IoU assignment in the second stage.

For ResNet-50 system-level comparisons, we additionally use 9 transformer encoder layers and evaluate with 300 predictions.
In our ablations, we use the default Deformable-DETR~\cite{zhu2020deformable} setting with 300 queries, 4 feature scales, 6 transformer encoder and decoder layers.
All other training hyper-parameters strictly follow Deformable-DETR~\cite{zhu2020deformable},
i.e., we use ResNet50 backbone with AdamW optimizer with learning rate $0.002$, and batch size $16$.
We train with a $1\times$ schedule, with 90K iteration and learning rate dropped at the 75K iteration.
Training takes $\sim\!18$ hours on 8 Quadro RTX 6000. 

\begin{table}
\begin{tabular}{lcc}
\toprule
Module              & \name& \ddetr{}  \\
\midrule                    
Backbone            & 22   & 22        \\
Transformer Encoder & 56   & 56        \\
Query Selection     & 13   & 9         \\
Transformer Decoder & 8    & 8         \\
Postprocess (NMS)   & 1    & 0         \\
\midrule                    
Total               & 100  & 95       \\
\bottomrule
\end{tabular}
\caption{\textbf{Runtime breakdown in milliseconds.}
Models use the ResNet-50 backbone.
The module names are taken from \reffig{pipeline}.
NMS in query selection and postprocessing takes 5ms.
}
\lbltbl{runtime_breakdown}
\end{table}

\subsection{Main results}
We compare Hungarian matching from the original DETR~\cite{carion2020end} with our IoU assignments under multiple datasets and architectures in \reftbl{ablation}.
We evaluate with Deformable DETR architecture on COCO under the $1\times$ schedule, DETR architecture on COCO under the $50$ epoch schedule and LVIS under 180k iterations.

We first report the results of the original 2-stage Deformable DETR~\cite{zhu2020deformable} alongside our improved baselines as described in \refsec{details}.
From the improved baseline, simply replacing the Hungarian matching with the naive traditional IoU assignments (without the object balance technique in \refsec{balance}) with NMS improves by $1.0$ mAP on COCO.
The improvements come with no additional cost except for NMS, which runs in a few milliseconds.
This shows traditional overlapping assignments can likewise train transformers.

However, with assignment alone, there is no change in LVIS AP and DETR's COCO AP.
On LVIS, traditional assignment underperforms on small object AP (21.5 vs 22.1) while outperforming on large object AP (50.0 vs 48.3).
We also see this trend in DETR where small object AP drops from 22.3 to 21.3 and large object AP increases from 59.7 to 61.2 on COCO.
We hypothesize traditional assignments cause an imbalance in the number of positives per object as the larger objects have many more matches (and appear less frequently in LVIS~\cite{gupta2019lvis}).
Further balancing the number of matched ground truth brings an additional gain on all settings.
On LVIS, object balancing increases AP by $2.4$ points.
On COCO, object balancing improves performance by $1.5,2.1$ mAP for Deformable DETR and vanilla DETR respectively.
The improvements are more pronounced for small objects on COCO, with a delta of $1.3$ mAP vs. $0.1$ mAP in large objects for Deformable DETR.

\begin{table*}[ht]
\centering
\begin{tabular}{@{}l@{\ }c@{\ \ \ \ }c@{\ \ \ \ }c@{\ \ \ \ }c@{\ \ \ \ }c@{\ \ \ \ }c@{\ \ \ \ }c@{\ \ \ }c@{\ \ \ }c@{}}
\toprule
Method       & Epochs & $AP$  & $AP_{50}$ & $AP_{75}$ & $AP_S$ & $AP_M$ & $AP_L$ & Params & FPS \\
\midrule
Faster RCNN~\cite{ren2015faster} & 12 & 37.9 & 58.8 & 41.0 & 22.4 & 41.1& 49.1 & 42M & \textit{17} \\
FCOS~\cite{tian2019fcos} & 12 & 38.6 & 57.4 & 41.4 & 22.3 & 42.5 & 49.8 & 32M & \textit{46} \\
ATSS~\cite{zhang2020bridging} & 12 & 39.3 & 57.5 & 42.8 & 24.3 & 43.3 & 51.3 & 32M & - \\  %
GFL~\cite{li2020generalizedv2} & 12 & 41.1 & 58.8 & 44.9 & 23.5 & 44.9 & 53.3 & 32M & \textit{19} \\
Cascade RCNN~\cite{cai2018cascade} & 12 & 42.1 & 59.8 & 45.8 & 24.3 & 45.2 & 54.8 & 72M & \textit{7} \\
DyHead~\cite{dai2021dynamic} & 12 & 42.6 & 60.1 & 46.4 & 26.1 & 46.8 & 56.0 & 39M & -\\
CenterNet2~\cite{zhou2021probablistic} & 12 & 42.9 &  59.5 & 47.0 & 24.1 & 47.0 & 56.2 & 72M & \textit{18} \\
HTC~\cite{chen2019hybrid} & 20 & 43.2 & 59.4 & 40.7 & 20.3 & 40.9 & 52.3 & 80M & \textit{3} \\
\midrule
DETR~\cite{carion2020end} & 500 & 43.3 & 63.1 & 45.9 & 22.5 & 47.3 & 61.1 & 41M & 37\\
Anchor DETR~\cite{wang2021anchor} & 50 & 44.2 & 64.7 & 47.5 & 24.7 & 48.2 & 60.6 & 39M & 17 \\
TSP-RCNN+~\cite{sun2020rethinking} & 96 & 45.0 & 64.5 & 49.6 & 29.7 & 47.7 & 58.0 & - & \textit{15} \\
Conditional DETR~\cite{meng2021conditional} & 108 & 45.1 & 65.4 & 48.5 & 25.3 & 49.0 & 62.2 & 44M & 34 \\
Efficient DETR~\cite{yao2021efficient} & 36 & 45.1 & 63.1 & 49.1 & 28.3 & 48.4 & 59.0 & - & -\\
Deformable-DETR~\cite{zhu2020deformable} & 50 & 46.2 & 65.2 & 50.0 & 28.8 & 49.2 & 61.7 & 40M & 18 \\
FQDet~\cite{picron2022fqdet} & 24 & 46.2 & 65.3 & 49.2 & 29.1 & 50.3 & 61.0 & 42M & - \\
DAB-Deformable-DETR~\cite{liu2022dab} & 50 & 46.9 & 66.0 & 50.8 & 30.1 & 50.4 & 62.5 & 47M & 15 \\
Dynamic DETR~\cite{dai2021dynamic} & 36 & 47.2 & - & - & - & - & - & 58M & - \\
DN-Deformable-DETR~\cite{li2022dn} & 50 & 48.6 & 67.4 & 52.7 & 31.0 & 52.0 & 63.7 & 48M & 6 \\
Group-DETR~\cite{chen2022group} & 12 & 49.8 & - & - & 32.4 & 53.0 & 64.2 & - & - \\
$\mathcal{H}$-Deformable-DETR~\cite{jia2022detrs} & 36 & 50.0 & 68.3 & 54.4 & 32.9 & 52.7 & 65.3 & 48M & 12 \\
DINO$^{\dagger}$ ~\cite{zhang2022dino} ($1\times$ schedule) & 12 & 49.4 & 66.9 & 53.8 & 32.3 & 52.5 & 63.9 & 47M & 5 \\
DINO$^{\dagger}$ ~\cite{zhang2022dino} ($2\times$ schedule) & 24 & 51.3 & \textbf{69.1} & 56.0 & \textbf{34.5} & 54.2 & 65.8 & 47M & 5\\ %
\midrule
Improved \ddetr{}~\cite{zhu2020deformable}  & 12 & 47.7 & 65.9 & 52.0 & 31.2 & 51.0 & 61.8 & 48M & 13 \\
Improved \ddetr{}~\cite{zhu2020deformable}  & 50 & 49.6 & 67.8 & 54.2 & 32.5 & 52.7 & 63.9 & 48M & 13 \\
Improved \ddetr{}$^{\ast}$~\cite{zhu2020deformable}  & 50 & 47.7 & 66.0 & 51.9 & 29.9 & 51.0 & 62.7 & 52M & 11 \\
Improved \ddetr{}$^{\dagger}$~\cite{zhu2020deformable}  & 50 & 49.8 & 68.3 & 54.4 & 33.2 & 52.9 & 64.0 & 48M & 13 \\
\name (Ours $1\times$ schedule)$^{\ast \dagger}$  & 12 & 50.5 & 67.6 & 55.3 & 33.1 & 54.7 & 65.2 & 52M & 10 \\
\name (Ours $2\times$ schedule)  & 24 & 51.1 & 68.5 & 56.4 & 34.3 & 55.4 & 65.4 & 48M & 13 \\
\name (Ours $2\times$ schedule)$^{\ast \dagger}$ & 24 & \textbf{51.6} & 69.0 & \textbf{56.7} & 34.0 & \textbf{55.8} & \textbf{66.5} & 52M & 10 \\

\bottomrule
\end{tabular}
\caption{
\textbf{System-level Comparisons to prior work on COCO validation with ResNet50 backbone.}
For each detector, we list the number of training epochs, detection accuracy, number of model parameters and frames-per-second (FPS).
FPS was measured on the same V100 machine whenever possible. Italic FPS were copied from the original publication.
Top block: traditional \nmsdets;
Middle block: transformer-based end-to-end object detectors.
Bottom block: our transformer-based and NMS-based detector.
\name enjoys the best of both worlds.
It trains as fast as NMS-based detectors and achieves performance competitive with transformer-based detectors.
Methods with $\dagger$ are evaluated with the top 300 predictions.
Methods with $\ast$ use 9 encoder layers.
}
\lbltbl{r50}
\end{table*}

\begin{table*}
\centering
\begin{tabular}{@{}l@{\ }l@{\ }c@{\ \ \ \ }c@{\ \ \ \ }c@{\ \ \ \ }c@{\ \ \ \ }c@{\ \ \ \ }c@{\ \ \ \ }c@{\ \ \ \ }c@{}}
\toprule
Method       & Backbone & Extra Data & $AP$  & $AP_{50}$ & $AP_{75}$ & $AP_S$ & $AP_M$ & $AP_L$ & FPS \\
\toprule
Deformable-DETR~\cite{zhu2020deformable} & ResNeXt-101-DCN & - & 50.1 & 69.7 & 54.6 & 30.6 & 52.8 & 65.6 & 4.6 \\
EfficientDet-D7x~\cite{tan2020efficientdet} & EfficientNet-D7x-BiFPN & - & 55.1 & {73.4} & 59.9 & - & - & - & \textit{6.5} \\
ScaledYOLOv4~\cite{wang2020scaled} & CSPDarkNet-P7 & - & 55.4 & {73.3} & 60.7 & {38.1} & {59.5} & 67.4 & \textit{16}\\
CenterNet2~\cite{zhou2021probablistic} & Res2Net-101-DCN-BiFPN & - & 56.4 & 74.0 & 61.6 & 38.7 & 59.7 & 68.6 & \textit{5} \\
CopyPaste~\cite{ghiasi2021simple} & EfficientNet-B7 & Objects365 &  56.0 & - & - & - & - & - & - \\
HTC++~\cite{chen2019hybrid, liu2021swin} & Swin-L    & - & 57.7 & - & - & - &  - & - & - \\
\midrule
$\mathcal{H}$-Deformable-DETR~\cite{jia2022detrs}& Swin-L & - & 58.3 & \textbf{77.1} & 63.9 & \textbf{39.8} & 61.5 & 72.7 & 4.6 \\
DINO~\cite{zhang2022dino} & Swin-L & - & \textbf{58.6} & 76.9 & 64.1 & 39.4 & 61.6 & 73.2 & 2.7 \\
Improved Deformable-DETR  & Swin-L & - & 56.6 & 75.6 & 61.9 & 38.8 & 60.4 & 73.5 & 4.3 \\
\name (Ours)  & Swin-L & - & 58.5 & 76.5 & \textbf{64.4} & 38.5 & \textbf{62.6} & \textbf{73.8} & 4.2 \\
\midrule
DINO~\cite{zhang2022dino} & Swin-L & Objects365 & 63.3 & - & - & - & - & - & 2.7 \\
\name (Ours)  & Swin-L & Objects365 & \textbf{63.5} & \textbf{80.4} & \textbf{70.2} & \textbf{46.1} & \textbf{66.9} & \textbf{76.9} & 4.2 \\
\bottomrule
\end{tabular}
\caption{
\textbf{Comparisons to prior work with larger backbones on COCO test-dev.}
The results are taken from original works except $\mathcal{H}$-Deformable-DETR and DINO where models are taken from their repositories and evaluated on the official test server.
We train our model with Swin-L backbone~\cite{liu2021swin} with a 2$\times$ schedule.
FPS are reported on the same V100 machine whenever possible.
Italicized FPS are taken from original texts.
Top block: traditional NMS-based detectors; Middle block: transformer-based end-to-end object detectors. Bottom block: transformer-based detectors with Objects365 pretraining. 
}
\lbltbl{swin}
\end{table*}

\subsection{Comparisons to prior works}
\lblsec{prior_works}

\reftbl{r50} compares \name to traditional \nmsdets{} and transformer-based end-to-end detectors on the COCO 2017 validation set with ResNet50 backbone.
Traditional \nmsdets{} are usually efficient to train and converge with a short training schedule (12 epochs).
Transformer-based end-to-end detectors require longer training schedules, i.e., from 36 to 500 epochs, and can achieve higher overall performance.

On the standard $1\times$ schedule, our model achieves the highest performance of 50.5 mAP, outperforming the next best method by 0.7 mAP.
On the longer $2\times$ schedule, our model sees a 0.3 mAP performance gain.
This suggests that traditional training assignments are competitive with bipartite-like matching in detection transformers and may even converge faster.
At inference, our model runs at a comparable speed to Deformable-DETR~\cite{zhu2020deformable}.
Specifically, it runs 5\% slower due to NMS taking 5 milliseconds.
A detailed runtime breakdown between Deformable-DETR and \name is shown in ~\reftbl{runtime_breakdown}.
DINO is slower partially because it uses larger feature resolutions, starting at $\frac{H}{4} \times \frac{W}{4}$.

\subsection{Comparisons on large backbones}
\lblsec{larger_backbones}
\reftbl{swin} compares our models to other detectors on larger backbones on COCO test-dev.
We train \name with Swin-L backbone for $2\times$ schedule on COCO with no additional data and obtain 58.5 mAP.
With the exception of backbone and training on 16 GPUs for $2\times$ schedule, this setup is identical to our setup in \reftbl{ablation}.
Our method outperforms our Deformable-DETR baseline by 1.9mAP and performs comparably to the latest end-to-end detection transformers such as $\mathcal{H}$-Deformable-DETR~\cite{jia2022detrs} and DINO~\cite{zhang2022dino}.

Additionally, we pre-train \name on Object-365~\cite{gao2019objects365} for $2\times$ schedule and finetune on COCO with a lower learning rate and larger image sizes for $2\times$ schedule.
We evaluate with single-scale testing with no test-time augmentations.
This achieves 63.5AP on the COCO test-dev set, compared to DINO's 63.3AP.
This verifies bipartite matching is not necessary on larger backbones.

\subsection{Analysis}
We analyze 3 key advantages of our IoU assignment over the one-to-one matching in the vanilla Deformable DETR.

\myparagraph{Our decoder learns an easier function.}
Intuitively, classifying and refining the boundary of a region is an easier task than global recognition and deduplication.
We quantify an "easier" task as one which trains well with lower model complexity and measure the performance drop when decreasing the number of transformer layers.
\reftbl{depth} studies the model performance with fewer encoder or decoder layers.

The results show \name is more robust to fewer encoder and decoder layers than \ddetr.
\ddetr degrades severely with 3 and 1 encoder layer(s), dropping by $2.0$ and $4.8$AP compared to $0.3$AP and $3.2$AP.
Interestingly, \ddetr degrades more significantly with fewer decoder layers.
When we evaluate our standard 6 encoder layer model with a single decoder layer, performance drops by $1.9$AP and $14.3$ AP for \name and \ddetr.
This supports that our model learns a simpler decoding process that can be expressed in fewer decoder layers.
This simpler decoding will be beneficial for future lightweight detectors.

\begin{table}[t]
\newcommand{\badnum}[2]{#1 \textcolor{scarletred3}{\scriptsize{(-#2)}}}
\centering
\begin{tabular}{@{}c@{\ \ }c@{\ }c@{\ \ \ }c@{}}
\toprule
\# Enc. layer & \# Dec. layer & \name & \ddetr \\
\midrule
6  & 6 & 46.5 & 43.2 \\
\midrule
\multirow{2}{*}{6}  & 3 & \badnum{46.3}{0.2} & \badnum{42.5}{0.7} \\
 & 1 & \badnum{44.6}{1.9} & \badnum{28.9}{14.3} \\
\midrule
  & 6 & \badnum{46.2}{0.3} & \badnum{41.2}{2.0} \\
3  & 3 & \badnum{43.3}{3.2} & \badnum{38.6}{4.6} \\
 & 1 & \badnum{41.3}{5.2} & \badnum{24.1}{19.1} \\
\midrule
  & 6 & \badnum{43.3}{3.2} & \badnum{38.4}{4.8} \\
1  & 3 & \badnum{40.3}{6.2} & \badnum{34.3}{4.1} \\
  & 1 & \badnum{36.6}{9.9} & \badnum{18.2}{25.0} \\
\bottomrule
\end{tabular}
\caption{\textbf{Number of transformer layers.}  We experiment with ResNet50 and $1\times$ schedule without our improved baseline and object balancing technique. We report AP. The top row is the default setting. \ddetr performance drops much more than \name when using fewer layers. We verify \textbf{our decoder learns an easier function}.
}
\lbltbl{depth}
\vspace{-5mm}
\end{table}

\myparagraph{More positive samples help.}
\lblsec{inefficiencies}
Our IoU assignments inherently and implicitly assign one or more positive queries to each ground truth, while one-to-one matching guarantees a single positive.
We hypothesize more positive samples can speed up training and improve performance~\cite{li2022dn}.
To verify this hypothesis, we design a one-to-\textbf{many} matching \ddetr model that matches each ground truth to \emph{exactly} $B$ queries.
This is typically referred to as B-matching and can be simply reduced to bipartite matching by duplicating the cost matrix $B$ times.
As this objective encourages overlapping predictions, we add NMS.
\reftbl{bipartite} show the results of this \ddetr variant.
Surprisingly, we show this simple (and suboptimal) variant brings non-trivial improvements ($+2.5$ mAP under proper hyper-parameters).
This suggests more positive samples are important in detection.
Note that this matching scheme is balanced by design, with a fixed number of predictions ($B$) for each object.

\myparagraph{\name does not need self-attention in the decoder.}
We study the role of self-attention and cross-attention in the transformer decoder in~\reftbl{attention}.
We independently replace an attention mechanism with a feed-forward network to maintain parameter parity.

Replacing cross-attention is harmful to both \name and \ddetr{}.
We postulate that heavy MLP processing of queries harms performance.
Similar conclusions have been made about two-stage detectors, evident by their shallow region-of-interest heads.

Replacing self-attention is detrimental to \ddetr{} only.
\ddetr{} outputs unique objects by optimizing a set objective.
Thus, it is natural to require the self-attention set operation during decoding.
Meanwhile, \name can decode queries independently \footnote{An astute observer will notice queries can interact via cross attention. We clip deformable sampling to the boxes without performance drop.}.

\subsection{Ablation on object balancing}
\lblsec{positives}
An important design of \name is to balance the number of predictions for each object during training.
\reftbl{balance} varies $K$, the max number of predictions for each ground truth object.
A small $K$ will aggressively filter assigned predictions and may slow down training.
A large $K$ may ignore detrimental sampling biases.
$K=\infty$ does not balance objects.

The performance sweet spot at $K=4$ improves small, medium, large AP by $1.6$, $1.3$, $0.5$ over no balancing.
Notably, the gain comes from smaller object performance.
This is likely caused by a sampling bias resulting in fewer small object queries.
Surprisingly, we observe a performance gap between $K=1$ and bipartite matching despite both methods matching ground truth objects once.
$K=1$ improves small, medium, large AP by $2.7$, $1.3$, $1.4$ points, respectively.
This indicates that the sampling mechanism in vanilla bipartite matching is suboptimal, especially on small objects.

\begin{table}
\centering
\begin{tabular}{@{}c@{\ }c@{\ \ \ }c@{\ \ \ \ \ \ }c@{\ \ \ \ \ \ }c@{\ \ \ \ \ \ }c@{}}
\toprule
  First-stage & Second-stage & $AP$ & $AP_S$ & $AP_M$ & $AP_L$ \\
\midrule
1 & 1 & 43.2 & 61.9 & 46.7 & 25.9 \\
\midrule
1 & 2 & 44.7 & 64.1 & 48.0 & 26.8 \\
1 & 4 & 45.1 & 64.0 & 48.5 & 27.1 \\
1 & 8 & 43.0 & 62.2 & 46.0 & 24.5 \\
\midrule
2 & 1 & 44.0 & 62.8 & 47.4 & 27.7 \\
4 & 1 & 43.7 & 62.2 & 47.5 & 25.6 \\
8 & 1 & 43.1 & 61.3 & 46.8 & 26.4 \\
\midrule
2 & 2 & 45.4 & \textbf{64.7} & 48.9 & \textbf{28.6} \\
4 & 4 & \textbf{45.8} & 64.6 & \textbf{49.5} & 28.4 \\
\bottomrule
\end{tabular}
\caption{\textbf{More positive matchings to \ddetr{} help.}  We experiment with ResNet50 and $1\times$ schedule without our improved baseline. Top row: the default setting. Adding some positives to both the first and second stage helps.
}
\lbltbl{bipartite}
\end{table}

\begin{table}[t]
\centering
\begin{tabular}{@{}c@{\ \ \ \ \ \ }c@{\ \ \ \ \ \ }c@{\ \ \ \ \ \ }c@{}}
\toprule
Cross-attn & Self-attn &  \name & \ddetr \\
\midrule
\checkmark & \checkmark & 46.5 & 43.2 \\
\midrule
\checkmark &            & 46.6 & 34.7 \\
          & \checkmark &  9.3 &  7.2 \\
\bottomrule
\end{tabular}
\caption{
\textbf{Analysis of attention layers in the transformer decoder.}
We report AP on COCO validation with ResNet50 and $1\times$ schedule without improved baseline or object balancing technique.
The top row is the default setting.
We replace all self-attention layers (2nd row) or cross-attention layers (3rd row) with MLPs for \name and \ddetr.
Both detectors need cross-attention, but \name does not need self-attention in the decoder.
}
\lbltbl{attention}
\end{table}

\section{Limitations}
\name re-introduces NMS postprocessing to remove overlapping objects.
The NMS thresholds can be task-specific and challenging to tune.
For example, while NMS with a threshold of $0.5$ is fine for COCO images, it struggles with CrowdHuman~\cite{shao2018crowdhuman} images that contain objects in a crowd scene.
\name no longer specifies objects uniquely with queries.
This may change how the model will leverage queries in downstream tasks.
\name assumes we have an initial bounding box representation to which we can assign an object.
This is realized by the two-stage framework where the queries are initialized by a bounding box.
One drawback is that the method can only work with queries that represent bounding boxes.
However, most well-performing transformer-based detectors seem to use two-stage query formulation~\cite{zhang2022dino,zhu2020deformable}.

\begin{table}[!t]
\centering
\begin{tabular}{@{}l@{\ \ \ \ \ \ }c@{\ \ \ \ \ \ }c@{\ \ \ \ \ \ }c@{\ \ \ \ \ \ }c@{}}
\toprule
$K$  & $AP$ & $AP_S$ & $AP_M$ & $AP_L$\\
\midrule
$\infty$ & 46.5 & 28.4 & 50.2 & 63.0 \\
\midrule
 16 & 47.0 & 27.6 & 50.8 & 63.2 \\
  8 & 47.6 & 29.2 & 51.3 & \textbf{63.8} \\
  4 & \textbf{47.9} & \textbf{30.0} & \textbf{51.5} & 63.5 \\
  1 & 46.1 & 28.6 & 49.4 & 60.3 \\
\midrule
Bipartite  & 43.2 & 25.9 & 46.6 & 57.6 \\
\bottomrule
\end{tabular}
\caption{\textbf{Ablation on Object Balancing of \name in \refsec{balance}}.
We report results on COCO validation with ResNet50 and $1\times$ schedule without the improved baseline.
K is the hyper-parameter for the maximum predictions of a ground truth object.
$K=\infty$ (first row) means without object balancing.
We sweep $K$ in a log scale (middle block).
We also compare to the default one-to-one bipartite in Deformable DETR in the bottom row.
Object balancing improves AP by $1.4\%$, and is especially useful for small and medium objects.
}
\lbltbl{balance}
\vspace{-5mm}
\end{table}

\section{Conclusions}
Assignments effectively train detection transformers.
Our proposed training with IoU-based label assignment is more efficient than vanilla one-to-one matching of detection transformers.
We analyzed empirically that assignment-based training benefits from 1) learning easier decoding function, 2) learning from more positive samples, and 3) not needing to globally optimize via self-attention.
From our analyses, we showcase an arguably simpler alternative training mechanism of detection transformer compared to recent detection transformer methods~\cite{zhang2022dino, li2022dn}.
This alternative enjoys a significant advantage in training efficiency, especially with a short training schedule. 
We deduce that the effectiveness of detection transformer comes from its transformer encoder and decoder architecture.

\myparagraph{Acknowledgements} This material is in part based upon work supported by the National Science Foundation under Grant No. IIS-1845485 and IIS-2006820.
\definecolor{redvis}{RGB}{  164, 0,   0}
\definecolor{greenvis}{RGB}{  62, 123,   4}
\definecolor{bluevis}{RGB}{  32,  74, 135 }
\begin{figure*}[!t]
    \centering
    \small
    \rotatebox{90}{\quad \space \space \quad \name (Ours) \qquad \ddetr{} \qquad \quad Ground Truth}
    \includegraphics[width=2.0\columnwidth]{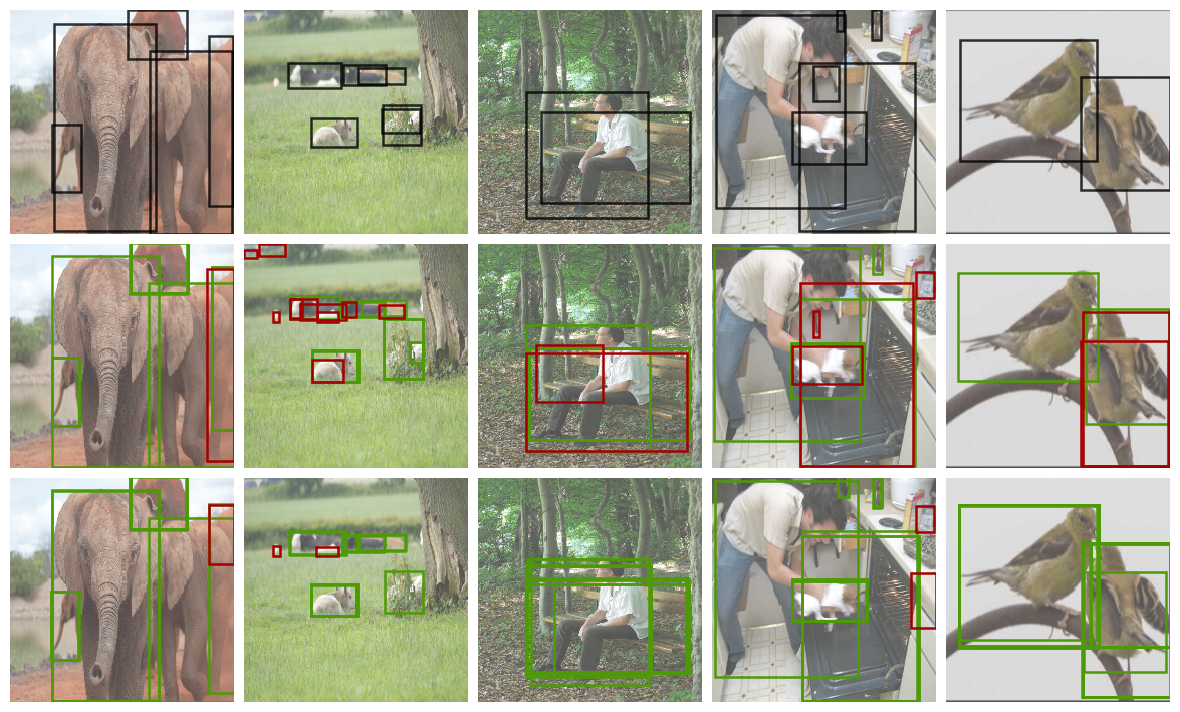}      %
    \caption{{\textbf{
    Example visualization of \textcolor{greenvis}{positive} and \textcolor{redvis}{negative} matches on COCO validation.
    }
    We visualize images with ground truth boxes (top), \ddetr{} predictions (middle), and our predictions (bottom).
    We use the oracle labels only to match each ground truth object.
    Our prediction color scheme is as follows.
    \textcolor{greenvis}{Green}: prediction matched to a ground truth object.
    \textcolor{redvis}{Red}: high confidence (>0.3) prediction matched to background.
    We find that bipartite matching in \ddetr{} does not match well-localized object predictions. 
    }}
    \lblfig{qual}
\end{figure*}
{
\small
\bibliographystyle{ieee_fullname}
\bibliography{egbib}
}

\end{document}